\documentclass{article}
\usepackage{ijcai23}

\usepackage{times}
\usepackage{soul}
\usepackage{url}
\usepackage[hidelinks]{hyperref}
\usepackage[utf8]{inputenc}
\usepackage[small]{caption}
\usepackage{graphicx}
\usepackage{amsmath}
\usepackage{amsthm}

\usepackage{booktabs}
\usepackage{algorithm}
\usepackage{algorithmic}
\usepackage[switch]{lineno}
\usepackage{color}
\usepackage{multirow}
\usepackage{amssymb}
\usepackage{subfig}
\usepackage{graphicx}

\newcommand{\methodname}{\textsc{Hi-GAS}}

\title{Hierarchical Federated Learning Incentivization for Gas Usage Estimation}

\author{
Hao Sun$^{1*}$
\and
Xiaoli Tang$^{2*}$
\and
Chengyi Yang$^{1*}$
\and
Zhenpeng Yu$^{1}$
\and
Xiuli Wang$^{1}$
\and\\
Qijie Ding$^{1}$
\and
Zengxiang Li$^{1}$
\And
Han Yu$^{2}$
\affiliations
$^{1}$ENN Group, Beijing, China\\
$^{2}$School of Computer Science and Engineering, Nanyang Technological University, Singapore\\
\emails
\{xiaoli001, han.yu\}@ntu.edu.sg
}

\begin{document}

\maketitle

\begin{abstract}
Accurately estimating gas usage is essential for the efficient functioning of gas distribution networks and saving operational costs. Traditional methods rely on centralized data processing, which poses privacy risks. Federated learning (FL) offers a solution to this problem by enabling local data processing on each participant, such as gas companies and heating stations. However, local training and communication overhead may discourage gas companies and heating stations from actively participating in the FL training process.
To address this challenge, we propose a Hierarchical FL Incentive Mechanism for Gas Usage Estimation (\methodname{}), which has been testbedded in the ENN Group, one of the leading players in the natural gas and green energy industry. It is designed to support horizontal FL among gas companies, and vertical FL among each gas company and heating station within a hierarchical FL ecosystem, rewarding participants based on their contributions to FL.
In addition, a hierarchical FL model aggregation approach is also proposed to improve the gas usage estimation performance by aggregating models at different levels of the hierarchy. The incentive scheme employs a multi-dimensional contribution-aware reward distribution function that combines the evaluation of data quality and model contribution to incentivize both gas companies and heating stations within their jurisdiction while maintaining fairness. Results of extensive experiments validate the effectiveness of the proposed mechanism. 
\end{abstract}

\section{Introduction}
\label{sec:introduction}
\def\thefootnote{*}\footnotetext{These authors contributed equally to this work.}
Gas usage estimation is of paramount importance for energy companies like ENN Group\footnote{https://www.enn.cn/} as it enables them to accurately forecast and plan their gas purchase and distribution requirements. Accurate gas usage estimation ensures that the energy company can efficiently manage their gas distribution networks, avoid shortages or surpluses of gas supply, and ultimately minimize operational costs. Moreover, gas usage estimation is a critical component of the energy company's efforts to reduce their carbon footprint and meet their sustainability goals \cite{le2020temporary}. 
However, the success of traditional methods for gas usage estimation is heavily reliant on large volumes of high-quality data. However, data from a single company may not be sufficient to train models effectively since data are often collected and owned by different organizations within a given field. Collaborative model training \cite{warnat2021swarm,chen2023efficient} has been identified as a valuable technique to enhance the quality of ML solutions by leveraging the collective data resources of multiple organizations. Federated Learning (FL) is an important category of collaborative model training framework that has gained popularity due to its ability to protect data privacy and user confidentiality \cite{yang2019federated,Liu-et-al:2020FedVision,Liu-et-al:2022IAAI}.

FL operates by having data owners (referred to as FL clients) train a local model using their private data samples, after which they submit model parameters (not raw training data) to a remote server. Once sufficient parameters from local models have been collected, a global model is aggregated and distributed to data owners for the next round of local training. This iterative process continues until the global model meets the predefined accuracy requirements. Through this training process, FL significantly enhances the data privacy of data owners since raw data is not uploaded.

Despite the significant benefits of FL, it faces several critical challenges, making its further development and broader application in real-world industries challenging \cite{zeng2021comprehensive,tu2022incentive}. Firstly, the data owners or clients typically consume their own resources, such as computing and communication resources for local training. As a result, self-interested clients may not be willing to contribute their resources for FL model training unless they receive sufficient economic compensation. Secondly, some unreliable clients may engage in undesirable behavior, which can negatively impact the performance of the global model for an FL task. In particular, a client may maliciously disturb its data and send low-quality updates to mislead the global model parameters, resulting in the failure of collaborative learning. These factors have given rise to FL incentive mechanisms \cite{khan2020federated,zhan2021survey}, which can be defined as the process of identifying the most optimal payment and organizational structure for the federation to attain desired operational objectives. 


In this paper, we propose the Hierarchical Federated Learning Incentive Mechanism for Gas Usage Estimation (\methodname{}). 
\methodname{} is based on a hierarchical federated learning ecosystem composed of one horizontal FL and several vertical FL. The horizontal FL enables gas companies of ENN to leverage the historical gas supply information and weather data owned by others to make accurate gas usage estimations. On the other hand, the vertical FL is designed to facilitate collaboration between each gas company and the heating stations within their area of responsibility, taking into consideration that the data owned by each gas company and its associated heat stations are vertically-partitioned.
To incentivize active participation and ensure fairness among gas companies and heating stations, we incorporate a multi-dimensional contribution-aware reward distribution function that considers both data quality and model contributions to \methodname{}. This hierarchical incentive scheme has proven effective in motivating participation and improving overall performance.
\methodname{} has been successfully implemented in ENN Group in one province of China, and has allowed two gas companies in separate cities to improve their gas usage forecasting accuracy. It has been successful in motivating gas companies and heating stations to actively participate in FL training and commit high-quality data, resulting in increased revenue for these entities. To our best knowledge, it is the first successfully hierarchical federated learning incentive approach for the energy industry.


\section{Related Work}
In federated learning, incentive mechanisms typically involve addressing sub-problems such as contribution evaluation, node selection, and payment allocation, as highlighted in \cite{zeng2021comprehensive}. Of these, contribution evaluation is particularly relevant to our work, and we provide a brief survey of existing literature in this area.

Existing approaches for contribution evaluation in federated learning can be broadly divided into four categories: self-reporting, individual performance, utility game, and Shapley Value (SV)-based methods \cite{zeng2021comprehensive}.

Self-reporting approaches \cite{yu2020sustainable,zeng2020fmore,zhang2020hierarchically,ding2020incentive,feng2019joint} measure participants' contributions based on their self-reported information regarding their sensitive local data such as data quantity, quality, committed computational and communication resources. For example, \cite{yu2020sustainable} proposes an incentive mechanism to compensate participants for their contributions and costs for joining the federation, measured based on self-reported data quantity and quality. However, this approach suffers from the possibility of dishonest reporting, where participants may overstate their contribution to receive a higher reward. As such, this approach is not ideal for large-scale and complex federated learning scenarios.

Individual performance-based approaches \cite{zhao2021crowdsensing,lyu2020collaborative,pandey2020crowdsourcing} assign a contribution value to each participant based on their individual performance on specific tasks. For example, \cite{zhao2021crowdsensing} measures individual contributions based on the similarity between local model updates and the aggregated FL model. While these approaches have been successful, they do not consider the contributions of other participants, which may lead to unfair reward distribution. 

Utility game-based approaches rely on the changes in coalition utility when a participant joins the federation \cite{wang2019measure,ghorbani2019data,nishio2020estimation}. In this category, there are three profit-sharing principles: egalitarian, marginal gain-based, and marginal loss-based. Fair value game, labor union, and Shapley Value-based game are the most common profit-sharing schemes. These methods may face challenges in designing a utility function that accurately reflects the contribution of each participant.

SV-based approaches have been extensively researched in recent years due to their ability to calculate a participant's contribution fairly \cite{shapley1953value}. However, the original SV calculation can be computationally expensive due to its exponential nature. To improve efficiency, researchers have proposed various techniques such as random sampling Monte-Carlo (MC) estimation \cite{castro2009polynomial} and the use of the fisher Information Matrix \cite{tang2021incentive}. These approaches reduce the number of model trainings needed to calculate SV, which may not be practical for large-scale FL applications.

\section{The proposed \methodname{} Approach}
In this section, we will give a detailed description of the proposed \methodname{}, which is based on a hierarchical federated learning ecosystem and tries to fairly distribute rewards to participants in order to effectively motivate them actively join in the FL training, improving the gas usage estimation performance. 

\begin{figure*}[t]
\centering
\includegraphics[width=\linewidth]{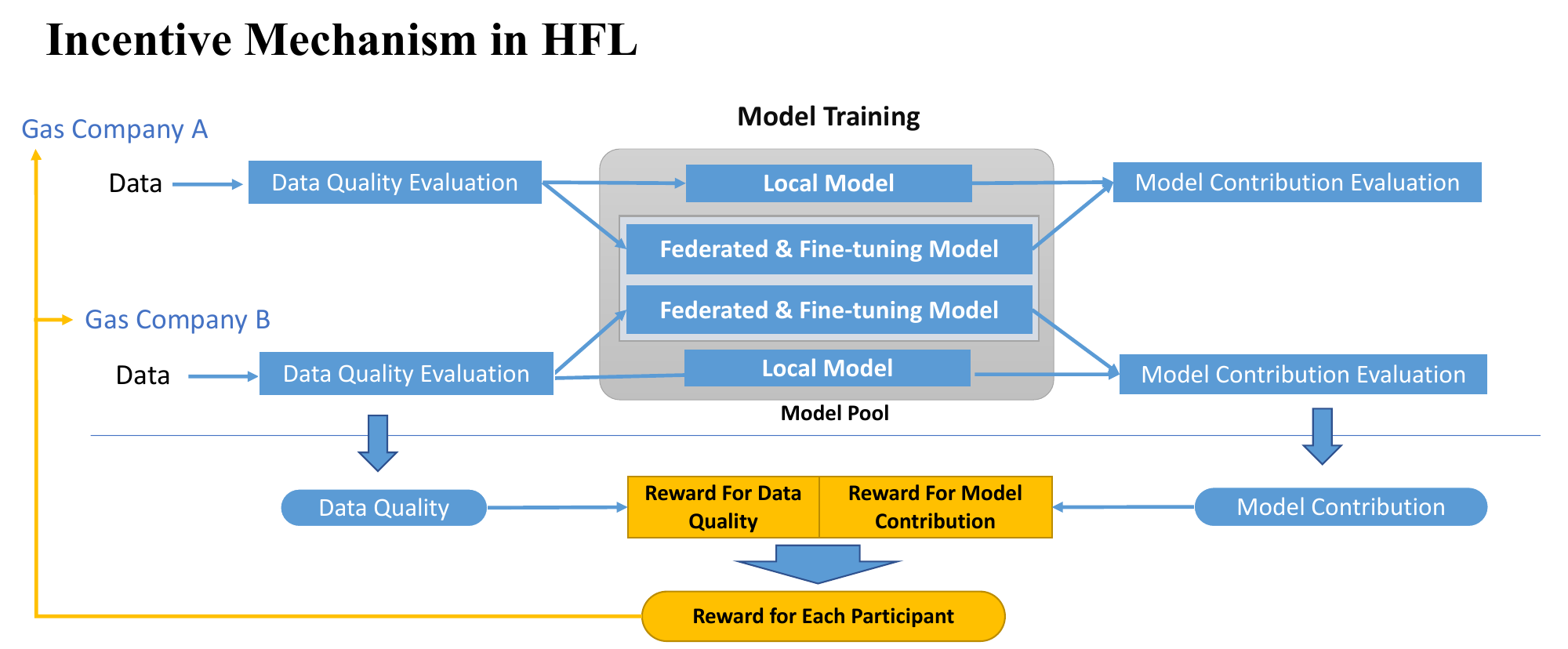}
\caption{Overview of the incentive model in \methodname{} for the HFL scenario.}
\label{fig:HFL}
\end{figure*}

\begin{figure*}[t]
\centering
\includegraphics[width=\linewidth]{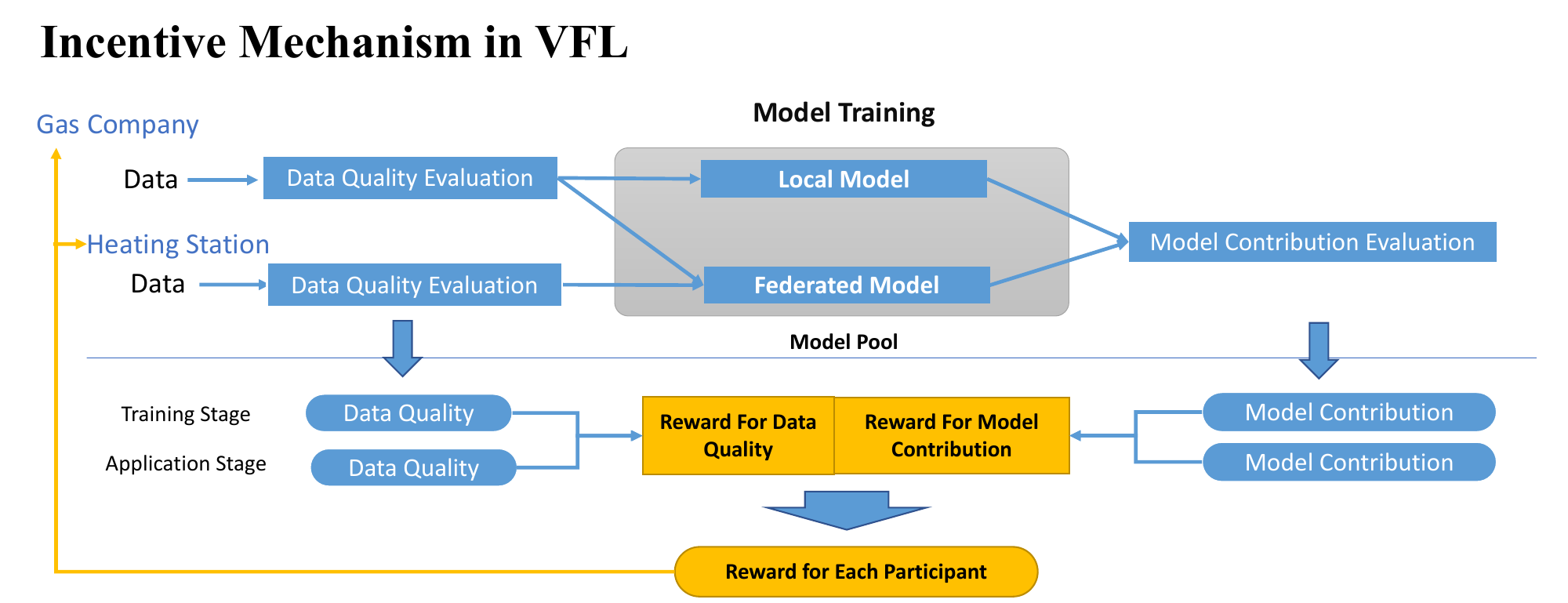}
\caption{Overview of the incentive model in \methodname{} for the VFL scenarios.}
\label{fig:VFL}
\end{figure*}

\subsection{Federated Learning under \methodname{}}
ENN Group's gas supply chain comprises two main participants: gas companies and heating stations. Gas companies purchase gas from external parties and then distribute it to heating stations within their jurisdiction. 
However, it is challenging for both gas companies and gas stations to make precise predictions based on their own data. Gas companies face the problem of data sparsity, which makes it difficult for them to train accurate gas demand prediction models using only their own data samples. Meanwhile, heating stations rely on heating strategists to develop daily heating plans and estimate gas usage based on manually-generated weather forecasts. However, the subjectivity involved in manually formulating strategies and the lack of high-precision weather forecast data available to heating stations can adversely affect the performance of these plans.
In this sense, our solution involves two federated learning ecosystems: a horizontal one (HFL) show in fig. \ref{fig:HFL} among gas companies and several vertical ones (VFL) shown in fig. \ref{fig:VFL} among each gas company and the heating stations within in the area that it is responsible for. 

In the client-server HFL system, suppose there are in general $n^H$ clients who can participate in FL-based gas usage estimation model training. Each client $i$ owns a local dataset $D_i^H =\{(\mathbf{x}^H_j, y^H_j)\}_{j=1}^{|D^H_i|}$. $\mathbf{x}^H_j$ denotes the $j$-th local training sample. $y^H_j$ denotes the corresponding ground truth label of $\mathbf{x}^H_j$. $|D^H_i|$ denotes the total number of data samples in $D^H_i$. The aim of HFL is to solve the following optimization problem under the aforementioned setting:
\begin{equation}
\label{eq:hfl_goal}
\begin{aligned}
\min_{\mathbf{\theta}^H}\sum_{i=1}^{n^H} \frac{|D^H_i|}{|D^H|} \mathcal{L}_i^H(\mathbf{\theta}^H;D^H_i),
\end{aligned}
\end{equation}
where $\mathbf{\theta}^H$ denotes the parameters of the model. $|D^H| = \sum_{i =1}^{n^H} |D^H_i|$ denotes the total number of samples. $\mathcal{L}_i^H(\mathbf{\theta}^H;D^H_i) = \frac{1}{|D^H_i|} \sum_{j=1}^{|D^H_i|}l(\mathbf{\theta}^H;\mathbf{x}^H_j, y^H_j)$ denotes the local loss of a given client $i$.

Different from HFL in which data are partitioned by sample, VFL assumes data are partitioned by feature. Let $n^V$ denote the number of participants in the VFL model training. A dataset $D^V = \{\mathbf{x}^V_j, y^V_j\}_{j=1}^{|D^V|}$ are partitioned across the $n^V$ participants. Each participant is associated with a unique set of features. Take it for example, the $i$-th block features $\mathbf{x}_{j,i}$ of the $j$-th sample $\mathbf{j} = [x_{j,1}^T, \cdots, x_{j,n^V}^T]^T$ are maintained by the $i$-th participant. 

\textbf{Active Participant}. The active participant is referred to the participant that holds not only data features but also labels of data samples. The active participant is the dominator during the VFL training since machine learning requires labels to derive the loss function \cite{xia2021vertical}.

\textbf{Passive Participant}. The passive participant is defined as the participant that only provides extra features during the VFL training but without labels of data samples \cite{xia2021vertical}.

Particularly, suppose the first participant is the active one, which means that the labels are partitioned to this participant. Following \cite{liu2022vertical,yang2023survey}, each participant $i$ trains the model parameter $\theta^V$ with its own local raw features $\mathbf{x}_{j,i}$ with the aim of minimizing the loss function as follows:
\begin{equation}
\label{eq:vfl_goal}
\begin{aligned}
\min_{\mathbf{\theta}^V} \frac{1}{n^V}\sum_{j=1}^{n^V}  \mathcal{L}^V(\mathbf{\theta}^V;D^V_j).
\end{aligned}
\end{equation}

\subsection{Multi-dimensional Contribution-aware Reward Distribution}
As show in fig. \ref{fig:HFL} and \ref{fig:VFL}, the incentive mechanism supports both the HFL among gas companies and the VFL among each gas company and the heating stations within its jurisdiction. The incentive mechanism is composed of four parts: data quality calculation, model contribution calculation, revenue allocation ration calculation, and reward for each participant calculation. In the following, we will present how each part works in detail. 

\subsubsection{Data Quality}
As depicted in figures \ref{fig:HFL} and \ref{fig:VFL}, the data quality evaluation model is utilized to preprocess the data of each participant before initiating the FL model training process. This model offers a data-centric approach to assess the quality of raw data, facilitating a precise estimation of the data quality.

Specifically, in the case of HFL, historical gas usage and weather conditions significantly impact the current gas usage. Thus, we evaluate the data quality of each participant by analyzing the correlation between their historical gas usage and weather data with actual gas usage data. If the correlation is strong, the quality of the historical gas usage and weather data is high; otherwise, it is considered unsatisfactory \cite{tahmasebi2012multiple}. Let $\mathbf{X}_i$ denote the historical gas usage and weather data of participant $i$ for a continuous period of $T$ days, and let $\mathbf{Y}_i$ represent the corresponding actual gas usage data. The correlation between $\mathbf{X}_i$ and $\mathbf{Y}_i$ can be formulated as:
\begin{equation}
\label{eq:data_correlation_hfl}
\begin{aligned}
\text{corr\_score}_i^H  = \frac{Cov(\mathbf{X}_i, \mathbf{Y}_i)}{\sqrt{Var(\mathbf{X}_i)Var(\mathbf{Y}_i)}},
\end{aligned}
\end{equation}
where the supscript $H$ means that the equation is applicable for the HFL setting. $Cov$ is the covariance function \cite{rice2006mathematical} and $Var$ is the variance function \cite{breiman2001random}.
Apart from the data quality, data quantity also reflect the quality of data. In this sense, in the HFL, we evaluate the quality of the data of each participant from these two perspectives. 
In specific, the data quantity value of participant $i$ is formulated as:
\begin{equation}
\label{eq:data_quantity_hfl}
\begin{aligned}
\text{quant\_score}_i^H = \frac{|D_i^H|}{|D^H|},
\end{aligned}
\end{equation}
where $|D_i^H|$ is the total number of samples of participant $i$. $|D^H|$ denotes the total number of samples in the HFL ecosystem. 

Then, we combine the data correlation score and the data quantity score to get the final evaluation result of the data quality of participant $i$ in the HFL setting as follows:
\begin{equation}
\label{eq:data_quality_hfl}
\begin{aligned}
\text{quality}_i^H = \text{corr\_score}_i^H \times \text{quant\_score}_i^H. 
\end{aligned}
\end{equation}

As mentioned previously, unlike the HFL approach where data is partitioned by samples, in the VFL setting, data is partitioned by features. Thus, we evaluate data quality of participant $i$ in VFL solely from the perspective of data correlation, which is defined as follows:
\begin{equation}
\label{eq:data_quality_vfl}
\begin{aligned}
\text{quality}_i^V = \text{corr\_score}_i^V, 
\end{aligned}
\end{equation}
where the definition of $\text{corr\_score}_i^V$ is the same as that of $\text{corr\_score}_i^H$. It is worth noting that, under the VFL setting, $\mathbf{X}_i$ represents either the historical heating strategies of heating stations or the weather information of the gas company but not both.

\subsubsection{Model Contribution}
Following existing incentive mechanisms in FL settings \cite{Liu-et-al:2022IAAI}, we also evaluate data from the perspective of model contribution. 

Specifically, to accurately assess the individual contribution of each participant in the federated learning process, each participant will train a local model exclusively on their own local data. These models are referred to as the Local Model as illustrated in figure \ref{fig:HFL} and \ref{fig:VFL}. Then, each participant predict the gas usage for $T$ consecutive days based on its local model and compare the results with those actual gas usage to calculate the Symmetric Mean Absolute Percentage Error (SMAPE) \cite{hyndman2006another}. SMAPE  is a commonly used evaluation metric in forecasting and time series analysis. It is used to measure the accuracy of a model's predictions by comparing the actual and predicted values of a time series.

The formula for SMAPE is as follows:
\begin{equation}
\label{eq:smape}
\begin{aligned}
\text{SMAPE} = \frac{1}{T} \sum_{t=1}^{T} \frac{|F_t - A_t|}{\frac{(|F_t| + |A_t|)}{2}},
\end{aligned}
\end{equation}
where $T$ is the number of time periods. $F_t$ is the forecasted value at time $t$. $A_t$ is the actual value at time $t$
The SMAPE metric measures the difference between the actual and predicted values, normalized by the average of the actual and predicted values. Unlike other percentage error metrics, SMAPE takes into account both the magnitude and direction of the error. Additionally, SMAPE is symmetric, meaning that overprediction and underprediction are weighted equally. The resulting SMAPE score is expressed as a decimal, with lower values indicating better accuracy. 

Due to the wide range of values (0\% to 200\%) produced by the SMAPE metric, we have modified it to ensure that its results always fall within the range of 0 to 100\%, for ease of subsequent calculations. We apply this modified variant of SMAPE to calculate the prediction error of each participant $i$'s local model as:
\begin{equation}
\label{eq:smape_new_local}
\begin{aligned}
\text{SMAPE\_new}^{local}_i = \frac{1}{T} \sum_{t=1}^{T} \frac{|y_{i,t}^{local} - \hat{y}_{i,t}|}{(|y_{i,t}^{local}| + |\hat{y}_{i,t}|)},
\end{aligned}
\end{equation}
where $\hat{y}_{i,t}$ represents the actual gas usage at time $t$ of participant $i$, and $y_{i,t}^{local}$ represents the prediction generated by the local model of participant $i$ at the same time $t$.

Similarly, we get the prediction error of the global model on gas usage of participant $i$':
\begin{equation}
\label{eq:smape_new_global}
\begin{aligned}
\text{SMAPE\_new}^{global}_i = \frac{1}{T} \sum_{t=1}^{T} \frac{|y_{i,t}^{global} - \hat{y}_{i,t}|}{(|y_{i,t}^{global}| + |\hat{y}_{i,t}|)},
\end{aligned}
\end{equation}
where $y_{i,t}^{global}$ is the predicted gas usage of participant $i$ at time $t$ generated by the global model.

Then, we get the accuracy of the local model and global model from the perspective of participate $i$ as
\begin{equation}
\label{eq:acc}
\begin{aligned}
\text{acc}_i^{local} = 1 - \text{SMAPE\_new}^{local}_i, \\
\text{acc}_i^{global} = 1 - \text{SMAPE\_new}^{global}_i.
\end{aligned}
\end{equation}

Based on $\text{acc}_i^{local}$ and $\text{acc}_i^{global}$, we can get the increment for participant $i$:
\begin{equation}
\label{eq:increment_i}
\begin{aligned}
\text{increment}_i = \text{acc}_i^{global} - \text{acc}_i^{local}.
\end{aligned}
\end{equation}
Here, $\text{increment}_i$ represents the benefit the participant $i$ got from the contribution of other participants by joining the FL training. 

Finally, we calculate the contribution of participant $j$ for the FL ecosystem as
\begin{equation}
\label{eq:contribution_j}
\begin{aligned}
\text{contribution}_j = \sum_{i \neq j} \frac{\text{acc}_i^{global} - \text{acc}_i^{local}}{n-1},
\end{aligned}
\end{equation}
where $n$ is the number of participants in the global model training. 

\subsubsection{Revenue Allocation Ratio}
To scale the data to a similar range and reduce bias, we normalize the quality and contribution of each participant $i$ as follows:
\begin{equation}
\label{eq:data_quality_norm}
\begin{aligned}
\text{quality}_i^{norm} = \frac{\text{quality}_i}{\sum_{j=1}^n \text{quality}_j}, 
\end{aligned}
\end{equation}
\begin{equation}
\label{eq:contribution_norm}
\begin{aligned}
\text{contribution}_i^{norm} = \frac{\text{contribution}_i}{\sum_{j=1}^n \text{contribution}_j}.
\end{aligned}
\end{equation}
Here, $n$ is the number of participants. 

\subsubsection{Reward for each participant}
Let $R_{data}$ and $R_{model}$ represent the total reward for data quality and model contribution, respectively. Then, the data quality reward and model contribution reward for participant $i$ are calculated as:
\begin{equation}
\label{eq:data_quality_reward}
\begin{aligned}
\text{r}_i^{quality} = R_{data} \times \text{quality}_i^{norm}, 
\end{aligned}
\end{equation}
\begin{equation}
\label{eq:contribution_reward}
\begin{aligned}
 \text{r}_i^{contribution} = R_{model} \times \text{contribution}_i^{norm}.
\end{aligned}
\end{equation}

\section{Experimental Evaluation}

In this section, we present results on testbedding \methodname{} in ENN Group energy plants across two cities.
Particularly, as the key contributions of this work are the hierarchical federated learning incentive mechanism and the multi-dimensional contribution-aware reward distribution mechanism, we conduct experiments to answer the following research questions:
\begin{itemize}
    \item \textbf{RQ 1}: Is the proposed incentive mechanism useful for motivating participants actively make commitment to the FL ecosystem?
    \item \textbf{RQ 2}: Whether the proposed method can comprehensively and accurately measure the contributions of all participants, and distribute rewards fairly based on their individual merits?
    \item \textbf{RQ 3}: Is it possible for the proposed multi-dimensional contribution-aware reward distribution mechanism to effectively evaluate the quality of data provided by participants?
\end{itemize}

In what follows, we will answer the above research questions one by one.


\subsection{Results and Discussion for RQ 1}
Table \ref{tab:improvement} presents a comparison of the data quantity, data quality value, and model contribution value for one gas company in the ENN Group before and after adopting the proposed \methodname{}. It is observed that the data quantity of the gas company decreases, while the data quality value increases significantly. This is because the gas company actively removed low-quality data when committing to the FL ecosystem after adopting \methodname{}. This indicates that the proposed \methodname{} effectively motivates participants to provide high-quality data.

Furthermore, as shown in Table \ref{tab:improvement}, the model contribution value of the gas company improves by 20.22\%. This can be attributed to the gas company being motivated to provide more high-quality data to the FL ecosystem, thereby contributing more to the improvement of the entire ecosystem. It is worth noting that the gas company receives a total reward increase after adopting \methodname{}. However, due to privacy concerns, we cannot disclose the exact results.
\begin{table}[t!]
\centering
\caption{Improvement by the proposed \methodname{} for a gas company. DQ, DQV, and MCV represent data quantity, data quality value and model contribution value, respectively.}
\begin{tabular}{|*{4}{c|}}
\hline
Gas company    & DQ & DQV & MCV \\\hline
B  & 11,414  &  0.0270 & 0.0925 \\\hline
B  & 9,404 &  0.8443 & 0.1112 \\\hline
\end{tabular}
\label{tab:improvement}
\end{table}

\subsection{Results and Discussion for RQ 2}
The data quantity, data quality value, and the reward allocation ratio based only on the data quality value of gas company A and B in the HFL ecosystem are shown in Table \ref{tab:data_quality_reward}. It is easy to see that the gas company with more data committed is with higher data quality value and gets higher reward allocation ratio, which means higher reward. In this sense, the proposed \methodname{} can effectively evaluate the contribution of each participant, based on which fairly distributes reward to each participant. 

\begin{table}[t!]
\centering
\caption{Comparison of data quality and corresponding rewards between two gas companies under \methodname{}. DQ, DQV, and RAR-DQV represent data quantity, data quality value and reward allocation ratio with only the data quality value, respectively.}
\begin{tabular}{|*{4}{c|}}
\hline
Gas company   & DQ & DQV & RAR-DQV \\\hline
A  & 530  &  0.0459 & 0.0516 \\\hline
B  & 9,404 &  0.8443 & 0.9484\\\hline
\end{tabular}
\label{tab:data_quality_reward}
\end{table}

Table \ref{tab:model_contribution_reward} compares the two gas companies in terms of data quantity, model contribution value, and reward allocation ratio, calculated using Eq. \eqref{eq:contribution_norm}.  
Gas company A's reward allocation ratio increases from 0.0516, as shown in Table \ref{tab:data_quality_reward}, to 0.1844 in Table \ref{tab:model_contribution_reward}. This increase is attributed to the high quality of the data provided by gas company A. The high-quality data improves gas company A's model contribution value, which, in turn, leads to a higher final reward allocation ratio. This indicates that even with a smaller amount of data, participants can still receive higher rewards as long as the data quality is high.

\begin{table}[t!]
\centering
\caption{Comparison of model contribution and corresponding reward between two gas companies under \methodname{}. DQ, MCV, and RAR represent data quantity, model contribution value and reward allocation ratio, respectively.}
\begin{tabular}{|*{4}{c|}}
\hline
Gas company    & DQ & MCV & RAR \\\hline
A  & 530  &  0.0251 & 0.1844 \\\hline
B  & 9,404 &  0.1112 & 0.8156\\\hline
\end{tabular}
\label{tab:model_contribution_reward}
\end{table}

\subsection{Results and Discussion for RQ 3}
To assess the effectiveness of \methodname{} in motivating heating stations of the VFL ecosystem to truthfully report their heating strategies and provide high-quality data, we compared the data quality value and model contribution value generated by randomly reported strategies and truthfully reported strategies. The results are presented in Tables \ref{tab:data_quality_value_under_strategy} and \ref{tab:model_contribution_value_under_strategy}.

Our analysis revealed that regardless of whether based on historical or real-time data, the data quality values for truthfully reported strategies are much higher than those for strategies generated randomly based on experience. Additionally, the model contribution value for truthfully reported strategies is significantly higher than that for randomly generated strategies during both the training and inference phases. These results suggest that \methodname{} is effective in motivating participants to truthfully commit their data.
\begin{table}[t!]
\centering
\caption{Comparison of data quality value across different gas usage strategies. DQV represents the data quality value. }
\begin{tabular}{|*{3}{c|}}
\hline
Data Type   & Strategy &  DQV \\\hline
\multirow{2} * {Historical} &  Experience + Randomness & 0.0997\\\cline{2-3}
{}  &  Strategy &  0.5130 \\\hline
\multirow{2} * {Real-time} &  Experience + Randomness & 0.1461\\\cline{2-3}
{}  &  Strategy &  0.7922 \\\hline
\end{tabular}
\label{tab:data_quality_value_under_strategy}
\end{table}
\begin{table}[t!]
\centering
\caption{Comparison of model contribution value across different gas usage strategies. MCV represents the model contribution value.}
\begin{tabular}{|*{3}{c|}}
\hline
Phase   & Strategy &  MCV \\\hline
\multirow{2} * {Training} &  Experience + Randomness & 0.0039\\\cline{2-3}
{}  &  Strategy &  0.0556 \\\hline
\multirow{2} * {Application} &  Experience + Randomness & 0.0220\\\cline{2-3}
{}  &  Strategy &  0.0598 \\\hline
\end{tabular}
\label{tab:model_contribution_value_under_strategy}
\end{table}

In a nusthell, the proposed \methodname{} has been effective in motivating participants to contribute data and participate in federated learning, leading to the creation of higher-accuracy models and significant cost saving. 

\section{Conclusions and Future Work}
In this paper, we propose a Hierarchical FL Incentive Mechanism for Gas Usage Estimation, which we implemented in the ENN Group, a leading player in the natural gas and green energy industry. Our proposed mechanism involves a hierarchical FL ecosystem that includes horizontal FL among gas companies and vertical FL among each gas company and the heating stations within its jurisdiction. We also developed a hierarchical incentive scheme that rewards participants based on their contributions to FL.
The hierarchical aggregation approach enhances the gas usage estimation performance by aggregating models at different levels of the hierarchy. The incentive scheme employs a multi-dimensional contribution-aware reward distribution function that evaluates both data quality and model contribution to incentivize both gas companies and heating stations within their jurisdiction while ensuring fairness. Extensive experiment results validate the effectiveness of our proposed mechanism.

In the future, we will comprehensively evaluate the proposed \methodname{} on more larger quantity of industrial datasets from more perspectives. In addition, we plan to improve the robustness of the proposed incentive mechanism against malicious participants \cite{lyu2022privacy} and further enhance fairness \cite{shi2023towards}.


\bibliographystyle{named}
\bibliography{ijcai23}

\begin{thebibliography}{}

\bibitem[\protect\citeauthoryear{Breiman}{2001}]{breiman2001random}
Leo Breiman.
\newblock Random forests.
\newblock {\em Machine learning}, 45:5--32, 2001.

\bibitem[\protect\citeauthoryear{Castro \bgroup \em et al.\egroup
  }{2009}]{castro2009polynomial}
Javier Castro, Daniel G{\'o}mez, and Juan Tejada.
\newblock Polynomial calculation of the shapley value based on sampling.
\newblock {\em Computers \& Operations Research}, 36(5):1726--1730, 2009.

\bibitem[\protect\citeauthoryear{Chen \bgroup \em et al.\egroup
  }{2023}]{chen2023efficient}
Yuanyuan Chen, Zichen Chen, Sheng Guo, Yansong Zhao, Zelei Liu, Pengcheng Wu,
  Chengyi Yang, Zengxiang Li, and Han Yu.
\newblock Efficient training of large-scale industrial fault diagnostic models
  through federated opportunistic block dropout.
\newblock {\em arXiv preprint arXiv:2302.11485}, 2023.

\bibitem[\protect\citeauthoryear{Ding \bgroup \em et al.\egroup
  }{2020}]{ding2020incentive}
Ningning Ding, Zhixuan Fang, and Jianwei Huang.
\newblock Incentive mechanism design for federated learning with
  multi-dimensional private information.
\newblock In {\em 2020 18th International Symposium on Modeling and
  Optimization in Mobile, Ad Hoc, and Wireless Networks (WiOPT)}, pages 1--8.
  IEEE, 2020.

\bibitem[\protect\citeauthoryear{Feng and others}{2019}]{feng2019joint}
Shaohan Feng et~al.
\newblock Joint service pricing and cooperative relay communication for
  federated learning.
\newblock In {\em iThings}, pages 815--820, 2019.

\bibitem[\protect\citeauthoryear{Ghorbani and Zou}{2019}]{ghorbani2019data}
Amirata Ghorbani and James Zou.
\newblock Data shapley: Equitable valuation of data for machine learning.
\newblock In {\em International Conference on Machine Learning}, pages
  2242--2251. PMLR, 2019.

\bibitem[\protect\citeauthoryear{Hyndman and
  Koehler}{2006}]{hyndman2006another}
Rob~J Hyndman and Anne~B Koehler.
\newblock Another look at measures of forecast accuracy.
\newblock {\em International journal of forecasting}, 22(4):679--688, 2006.

\bibitem[\protect\citeauthoryear{Khan \bgroup \em et al.\egroup
  }{2020}]{khan2020federated}
Latif~U. Khan, Shashi~Raj Pandey, Nguyen~H. Tran, Walid Saad, Zhu Han, Minh
  N.~H. Nguyen, and Choong~Seon Hong.
\newblock Federated learning for edge networks: Resource optimization and
  incentive mechanism.
\newblock {\em IEEE Communications Magazine}, 58(10):88--93, 2020.

\bibitem[\protect\citeauthoryear{Le~Qu{\'e}r{\'e} \bgroup \em et al.\egroup
  }{2020}]{le2020temporary}
Corinne Le~Qu{\'e}r{\'e}, Robert~B Jackson, Matthew~W Jones, Adam~JP Smith, Sam
  Abernethy, Robbie~M Andrew, Anthony~J De-Gol, David~R Willis, Yuli Shan,
  Josep~G Canadell, et~al.
\newblock Temporary reduction in daily global co2 emissions during the covid-19
  forced confinement.
\newblock {\em Nature climate change}, 10(7):647--653, 2020.

\bibitem[\protect\citeauthoryear{Liu \bgroup \em et al.\egroup
  }{2020}]{Liu-et-al:2020FedVision}
Yang Liu, Anbu Huang, Yun Luo, He~Huang, Youzhi Liu, Yuanyuan Chen, Lican Feng,
  Tianjian Chen, Han Yu, and Qiang Yang.
\newblock Fedvision: An online visual object detection platform powered by
  federated learning.
\newblock In {\em IAAI}, pages 13172--13179, 2020.

\bibitem[\protect\citeauthoryear{Liu \bgroup \em et al.\egroup
  }{2022a}]{liu2022vertical}
Yang Liu, Yan Kang, Tianyuan Zou, Yanhong Pu, Yuanqin He, Xiaozhou Ye,
  Ye~Ouyang, Ya-Qin Zhang, and Qiang Yang.
\newblock Vertical federated learning.
\newblock {\em arXiv preprint arXiv:2211.12814}, 2022.

\bibitem[\protect\citeauthoryear{Liu \bgroup \em et al.\egroup
  }{2022b}]{Liu-et-al:2022IAAI}
Zelei Liu, Yuanyuan Chen, Yansong Zhao, Han Yu, Yang Liu, Renyi Bao, Jinpeng
  Jiang, Zaiqing Nie, Qian Xu, and Qiang Yang.
\newblock Contribution-aware federated learning for smart healthcare.
\newblock In {\em IAAI}, pages 12396--12404, 2022.

\bibitem[\protect\citeauthoryear{Lyu \bgroup \em et al.\egroup
  }{2020}]{lyu2020collaborative}
Lingjuan Lyu, Xinyi Xu, Qian Wang, and Han Yu.
\newblock Collaborative fairness in federated learning.
\newblock {\em Federated Learning: Privacy and Incentive}, pages 189--204,
  2020.

\bibitem[\protect\citeauthoryear{Lyu \bgroup \em et al.\egroup
  }{2022}]{lyu2022privacy}
Lingjuan Lyu, Han Yu, Xingjun Ma, Chen Chen, Lichao Sun, Jun Zhao, Qiang Yang,
  and S~Yu Philip.
\newblock Privacy and robustness in federated learning: Attacks and defenses.
\newblock {\em IEEE transactions on neural networks and learning systems},
  2022.

\bibitem[\protect\citeauthoryear{Nishio \bgroup \em et al.\egroup
  }{2020}]{nishio2020estimation}
Takayuki Nishio, Ryoichi Shinkuma, and Narayan~B Mandayam.
\newblock Estimation of individual device contributions for incentivizing
  federated learning.
\newblock In {\em 2020 IEEE Globecom Workshops (GC Wkshps}, pages 1--6. IEEE,
  2020.

\bibitem[\protect\citeauthoryear{Pandey \bgroup \em et al.\egroup
  }{2020}]{pandey2020crowdsourcing}
Shashi~Raj Pandey, Nguyen~H Tran, Mehdi Bennis, Yan~Kyaw Tun, Aunas Manzoor,
  and Choong~Seon Hong.
\newblock A crowdsourcing framework for on-device federated learning.
\newblock {\em IEEE Transactions on Wireless Communications}, 19(5):3241--3256,
  2020.

\bibitem[\protect\citeauthoryear{Rice}{2006}]{rice2006mathematical}
John~A Rice.
\newblock {\em Mathematical statistics and data analysis}.
\newblock Cengage Learning, 2006.

\bibitem[\protect\citeauthoryear{Shapley and others}{1997}]{shapley1953value}
Lloyd~S Shapley et~al.
\newblock A value for n-person games.
\newblock {\em Classics in game theory}, 69, 1997.

\bibitem[\protect\citeauthoryear{Shi \bgroup \em et al.\egroup
  }{2023}]{shi2023towards}
Yuxin Shi, Han Yu, and Cyril Leung.
\newblock Towards fairness-aware federated learning.
\newblock {\em IEEE Transactions on Neural Networks and Learning Systems},
  2023.

\bibitem[\protect\citeauthoryear{Tahmasebi \bgroup \em et al.\egroup
  }{2012}]{tahmasebi2012multiple}
Pejman Tahmasebi, Ardeshir Hezarkhani, and Muhammad Sahimi.
\newblock Multiple-point geostatistical modeling based on the cross-correlation
  functions.
\newblock {\em Computational Geosciences}, 16:779--797, 2012.

\bibitem[\protect\citeauthoryear{Tang and Wong}{2021}]{tang2021incentive}
Ming Tang and Vincent~WS Wong.
\newblock An incentive mechanism for cross-silo federated learning: A public
  goods perspective.
\newblock In {\em IEEE INFOCOM 2021-IEEE Conference on Computer
  Communications}, pages 1--10. IEEE, 2021.

\bibitem[\protect\citeauthoryear{Tu \bgroup \em et al.\egroup
  }{2022}]{tu2022incentive}
Xuezhen Tu, Kun Zhu, Nguyen~Cong Luong, Dusit Niyato, Yang Zhang, and Juan Li.
\newblock Incentive mechanisms for federated learning: From economic and game
  theoretic perspective.
\newblock {\em IEEE Transactions on Cognitive Communications and Networking},
  2022.

\bibitem[\protect\citeauthoryear{Wang \bgroup \em et al.\egroup
  }{2019}]{wang2019measure}
Guan Wang, Charlie~Xiaoqian Dang, and Ziye Zhou.
\newblock Measure contribution of participants in federated learning.
\newblock In {\em 2019 IEEE international conference on big data (Big Data)},
  pages 2597--2604. IEEE, 2019.

\bibitem[\protect\citeauthoryear{Warnat-Herresthal \bgroup \em et al.\egroup
  }{2021}]{warnat2021swarm}
Stefanie Warnat-Herresthal, Hartmut Schultze, Krishnaprasad~Lingadahalli
  Shastry, Sathyanarayanan Manamohan, Saikat Mukherjee, Vishesh Garg, Ravi
  Sarveswara, Kristian H{\"a}ndler, Peter Pickkers, N~Ahmad Aziz, et~al.
\newblock Swarm learning for decentralized and confidential clinical machine
  learning.
\newblock {\em Nature}, 594(7862):265--270, 2021.

\bibitem[\protect\citeauthoryear{Xia \bgroup \em et al.\egroup
  }{2021}]{xia2021vertical}
Wensheng Xia, Ying Li, Lan Zhang, Zhonghai Wu, and Xiaoyong Yuan.
\newblock A vertical federated learning framework for horizontally partitioned
  labels.
\newblock {\em arXiv preprint arXiv:2106.10056}, 2021.

\bibitem[\protect\citeauthoryear{Yang \bgroup \em et al.\egroup
  }{2019}]{yang2019federated}
Qiang Yang, Yang Liu, Tianjian Chen, and Yongxin Tong.
\newblock Federated machine learning: Concept and applications.
\newblock {\em ACM Transactions on Intelligent Systems and Technology},
  10(2):12:1--12:19, 2019.

\bibitem[\protect\citeauthoryear{Yang \bgroup \em et al.\egroup
  }{2023}]{yang2023survey}
Liu Yang, Di~Chai, Junxue Zhang, Yilun Jin, Leye Wang, Hao Liu, Han Tian, Qian
  Xu, and Kai Chen.
\newblock A survey on vertical federated learning: From a layered perspective.
\newblock {\em arXiv preprint arXiv:2304.01829}, 2023.

\bibitem[\protect\citeauthoryear{Yu \bgroup \em et al.\egroup
  }{2020}]{yu2020sustainable}
Han Yu, Zelei Liu, Yang Liu, Tianjian Chen, Mingshu Cong, Xi~Weng, Dusit
  Niyato, and Qiang Yang.
\newblock A sustainable incentive scheme for federated learning.
\newblock {\em IEEE Intelligent Systems}, 35(4):58--69, 2020.

\bibitem[\protect\citeauthoryear{Zeng \bgroup \em et al.\egroup
  }{2020}]{zeng2020fmore}
Rongfei Zeng, Shixun Zhang, Jiaqi Wang, and Xiaowen Chu.
\newblock Fmore: An incentive scheme of multi-dimensional auction for federated
  learning in {MEC}.
\newblock In {\em ICDCS}, pages 278--288, 2020.

\bibitem[\protect\citeauthoryear{Zeng \bgroup \em et al.\egroup
  }{2021}]{zeng2021comprehensive}
Rongfei Zeng, Chao Zeng, Xingwei Wang, Bo~Li, and Xiaowen Chu.
\newblock A comprehensive survey of incentive mechanism for federated learning.
\newblock {\em arXiv preprint arXiv:2106.15406}, 2021.

\bibitem[\protect\citeauthoryear{Zhan \bgroup \em et al.\egroup
  }{2021}]{zhan2021survey}
Yufeng Zhan, Jie Zhang, Zicong Hong, Leijie Wu, Peng Li, and Song Guo.
\newblock A survey of incentive mechanism design for federated learning.
\newblock {\em IEEE Transactions on Emerging Topics in Computing},
  10(2):1035--1044, 2021.

\bibitem[\protect\citeauthoryear{Zhang \bgroup \em et al.\egroup
  }{2020}]{zhang2020hierarchically}
Jingfeng Zhang, Cheng Li, Antonio Robles-Kelly, and Mohan Kankanhalli.
\newblock Hierarchically fair federated learning.
\newblock {\em arXiv preprint arXiv:2004.10386}, 2020.

\bibitem[\protect\citeauthoryear{Zhao \bgroup \em et al.\egroup
  }{2021}]{zhao2021crowdsensing}
Bowen Zhao, Ximeng Liu, and Wei-neng Chen.
\newblock When crowdsensing meets federated learning: Privacy-preserving mobile
  crowdsensing system.
\newblock {\em arXiv preprint arXiv:2102.10109}, 2021.

\end{thebibliography}

\end{document}